\documentclass[journal]{IEEEtran}

\usepackage{cite}
\usepackage{epsfig}
\usepackage{graphicx}
\usepackage{amsmath}
\usepackage{amssymb}
\usepackage{diagbox}
\usepackage{subfigure}
\usepackage{multicol}
\usepackage{multirow}

\begin{document}
\title{Wearable Travel Aid for Environment Perception and Navigation of Visually Impaired People}

\author{Jinqiang~Bai,
		Zhaoxiang~Liu,
		Yimin~Lin,
		Ye~Li,
		Shiguo~Lian,~\IEEEmembership{Member,~IEEE,}
		Dijun~Liu

\thanks{J. Bai is with  is with the School of Electronic Information Engineering, Beihang University, Beijing, 10083, China, e-mail: baijinqiang@buaa.edu.cn.}
\thanks{Z. Liu, Y. Lin, Y. Li, and S. Lian are with the AI Department, CloudMinds Technologies Inc., Beijing, 100102, China, e-mail: {robin.liu, anson.lin, yale.li, scott.lian}@cloudminds.com.}
\thanks{D. Liu is with China Academy of Telecommunication Technology, Beijing, 10083, China, e-mail: liudijun@datang.com.}
\thanks{Manuscript received xx, xx; revised xx, xx.}}

\maketitle

\begin{abstract}
This paper presents a wearable assistive device with the shape of a pair of eyeglasses that allows visually impaired people to navigate safely and quickly in unfamiliar environment, as well as perceive the complicated environment to automatically make decisions on the direction to move. The device uses a consumer Red, Green, Blue and Depth (RGB-D) camera and an Inertial Measurement Unit (IMU) to detect obstacles. As the device leverages the ground height continuity among adjacent image frames, it is able to segment the ground from obstacles accurately and rapidly. Based on the detected ground, the optimal walkable direction is computed and the user is then informed via converted beep sound. Moreover, by utilizing deep learning techniques, the device can semantically categorize the detected obstacles to improve the users' perception of surroundings. It combines a Convolutional Neural Network (CNN) deployed on a smartphone with a depth-image-based object detection to decide what the object type is and where the object is located, and then notifies the user of such information via speech. We evaluated the device's performance with different experiments in which 20 visually impaired people were asked to wear the device and move in an office, and found that they were able to avoid obstacle collisions and find the way in complicated scenarios.
\end{abstract}

\begin{IEEEkeywords}
Wearable assistive device, navigation, Convolutional Neural Network, object recognition.
\end{IEEEkeywords}

\section{Introduction}\label{sec:Intro}
\IEEEPARstart{A}{ccording} to the World Health Organization (WHO), there are about 285 million visually impaired people all over the world~\cite{1}. Visual impairment makes it challenging for them to perform autonomous navigation and environment perception in unfamiliar environments~\cite{2,3}. Visually impaired people usually asked white canes or guide dogs for help when detecting obstacles over the past years. However, they may still suffer injuries from hanging obstacles, such as scaffoldings and portable ladders, as white canes and guide dogs can only detect obstacles at heights up to their chests~\cite{4,5}. Recently, Electronic Travel Aids (ETAs)~\cite{6} utilizing advanced sensing techniques have greatly improved the travelling experience of visually impaired people. However, the ultrasonic sensor based ETAs are poor at obstacle identification due to the wide beam angle of ultrasonic sensors, and the laser sensors based ETAs are expensive, heavy and have high power consumption, which make them unsuitable for wearable applications~\cite{4,7}. Although vision based ETAs (e.g. monocular camera based~\cite{8}, stereo camera based~\cite{9,10} and Red, Green, Blue and Depth (RGB-D) camera based~\cite{11,12}) have been widely used for assisting visually impaired people to avoid obstacles and find obstacle-free paths, some problems still exist. For example, a chair may be considered as an obstacle when a blind person is looking for a seat, or no path will be found to enter a room with closed doors. Whereas suppose the object recognition technique be adopted, the visually impaired people can find the chair to sit or open the door to enter the room. Therefore, there is a need for developing new assistive devices to help visually impaired people in navigation and environment perception.

\begin{figure}
	\begin{center}
		\includegraphics[scale=0.25]{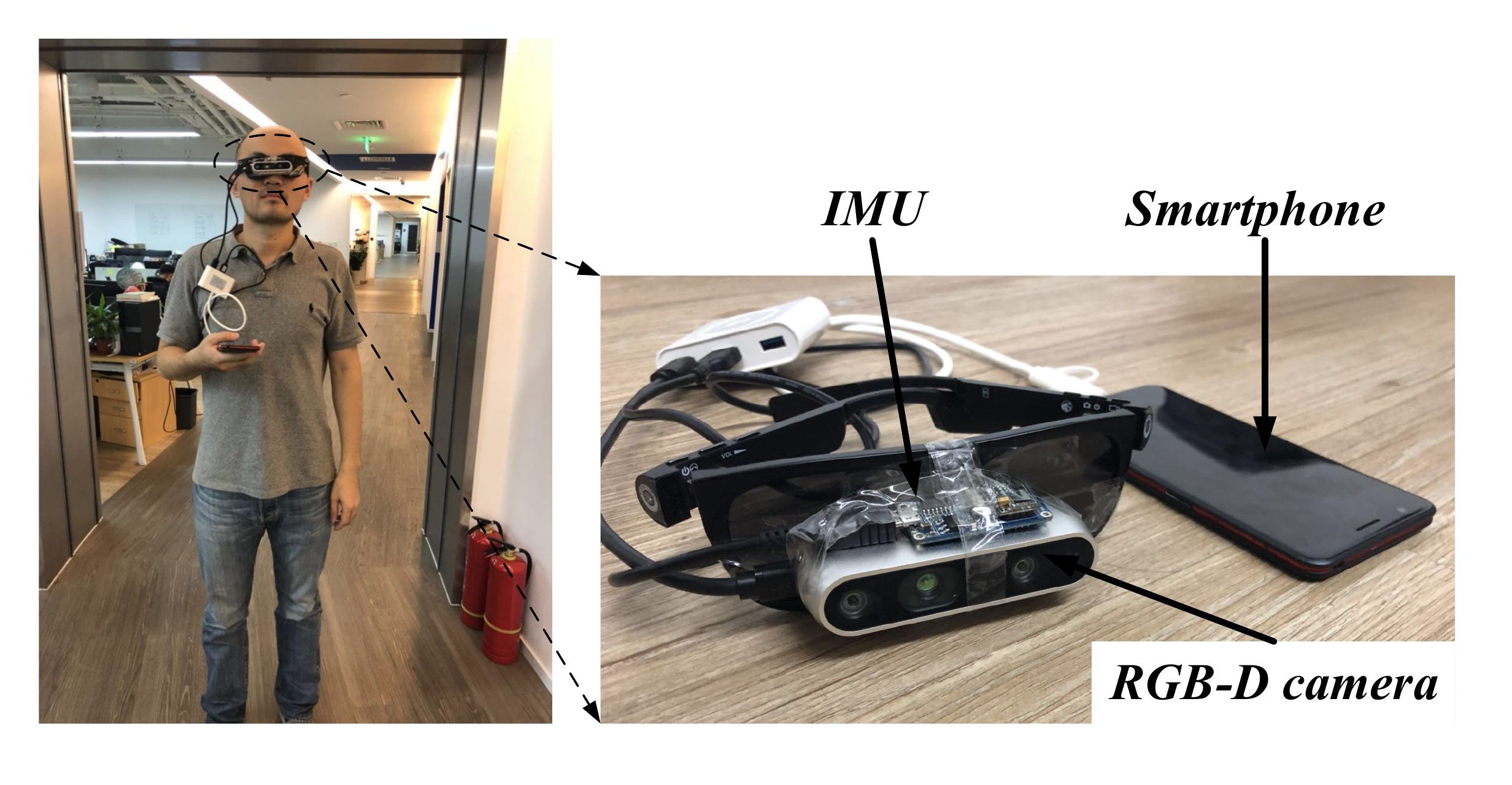}
	\end{center}
	\caption{The prototype of the proposed system.}
	\label{fig:1}
\end{figure}

This paper proposes a wearable assistive device (see Fig.~\ref{fig:1}) for visually impaired people's navigation and environment perception. The presented device detects the ground accurately and rapidly by leveraging an adaptive ground height segmentation algorithm and utilizing the ground height continuity among adjacent frames. With the detected ground, the optimal walkable direction can be computed and corresponding beep sound will be played out for blind navigation. Meanwhile, we adopt a lightweight Convolutional Neural Network (CNN) to semantically categorize the detected objects in the RGB image and then extract the object contour in the depth image to identify where the object is located. All the semantic information is finally converted to speech to inform the user about the perceived surroundings.

\section{Related Work}\label{sec:review}
This research builds on important related work of ground detection and object recognition. Because the vision based assistive devices have advantages over the ultrasonic or laser sensor based devices as mentioned above, this section focuses the relevant works on vision based ground detection and object recognition.

\subsection{Ground Detection}
General ground detection algorithms can be divided into scene segmentation based algorithms and surface normal vector estimation based algorithms.

\textbf{Segmentation:} Saitoh et al.~\cite{10} proposed a mean-shift algorithm to fit the ground plane. If the angle between the fitting ground and the horizontal plane is less than a threshold, the ground will be taken as a traversable area. However, the algorithm is too sensitive to the threshold. Rodriguez et al.~\cite{14} proposed a Random Sample Consensus (RANSAC)~\cite{15} based ground plane detection method and the potential obstacle was represented by polar grid. The ground detection error of this method reaches more than ten percent, leading to high obstacle detection errors. The wearability of the system and the integration of its main components into smaller devices should also be improved.

\textbf{Surface normal vector estimation:} Koester et al.~\cite{16} presented a gradient and surface normal vector based detection algorithm to compute the accessible sections effectively even in crowded scenes. However, the success rate of detection heavily relies on the quality of 3-Dimensional(D) reconstruction process. Bellone et al.~\cite{17} estimated the normal vectors to a local surface via Principal Component Analysis (PCA) and generated an unevenness point descriptor to detect traversable and non-traversable regions. The performance of the method was greatly affected by the search radius, which limits its applications in practice. Aladren et al.~\cite{7} used depth and color images to detect the ground in longer distance. Although their system achieved high ground detection accuracy, it is too computational expensive (2 frames per second) for visually impaired people to navigate in real time. Similar to our work, Imai et al.~\cite{18} proposed a ground detection approach which considered both ground height and normal vectors. However, since the ground height was computed only by one column data in depth image, it is prone to error if an obstacle happened to exist in this column. In contrast, the ground height in our approach is computed more robustly and rapidly through weighting the ground heights in previous and current frames.

\subsection{Object Recognition}
Tapu et al. \cite{21} presented a real-time obstacle detection and classification system for safe navigation of visually impaired people. The system used Scale Invariant Feature Transform (SIFT) and Features from Accelerated Segment Test (FAST) features to extract points of interest, and used Support Vector Machine (SVM) and Bag of Visual Words (BoVW) to classify these points. Although the system achieved 90\% precision of classification, the information about object distance is not available, and this will result in misguidance to blind individuals. Lee et al. \cite{11} proposed a robust depth-based obstacle detection system to obtain obstacle information which contains the distance, while it did not employ object recognition techniques, leading to poor environment perception functions.

Since AlexNet \cite{22} won the ImageNet Challenge: ILSVRC 2012 \cite{23}, CNN based object detection methods have become unprecedentedly popular. Although these methods have higher detection accuracy, they usually have large network sizes and high hardware performance requirements. Kaur et al. \cite{24} proposed a faster Recursive Convolutional Neural Network (RCNN) based scene perception system for visually impaired individuals. The system is able to provide the obstacle category and distance information. However, the distance information is obtained only through a single-line laser, making it unavailable or prone to error for some obstacles. Tapu et al. \cite{25} proposed a DEEP-SEE framework that used both computer vision algorithms and CNN to detect objects encountered during navigation. Although its recognition accuracy is satisfactory, the heavy computation load makes it difficult to be implemented on a smartphone.

Lightweight CNN based 2-D object detection algorithms \cite{26}\cite{27}\cite{28}\cite{29}\cite{30} have made great progress in recent years. They usually provide object category and location information in 2-D images, while there is still a lack of distance information. As a result, an object painted on the ground may be considered as an obstacle, and the visually impaired people will be misled.

To overcome the above limitations, we propose a 2.5-D object detection method that can provide the object category, distance and orientation information to make visually impaired people travel more easily.

\begin{figure}[b]
	\begin{center}
		\includegraphics[width=\linewidth]{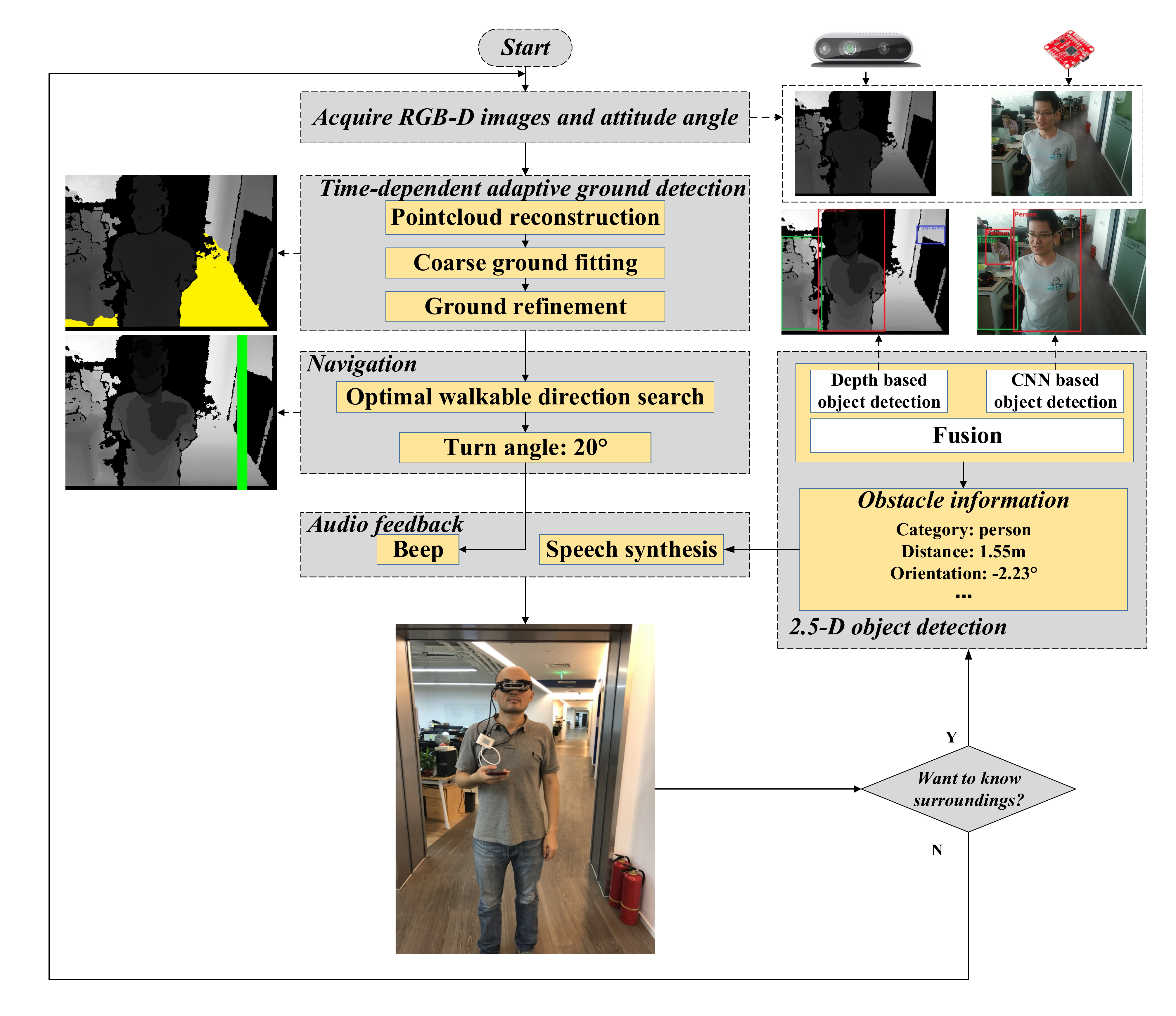}
	\end{center}
	\caption{The proposed system framework.}
	\label{fig:2}
\end{figure}

\section{System Design}\label{sec:system design}
As shown in Fig. \ref{fig:2}, the system first acquires RGB image and depth image from an RGB-D camera, and also obtains the camera attitude angle from an Inertial Measurement Unit (IMU) attached on the camera. Then based on the above sensor data, a time-dependent adaptive ground detection algorithm is performed to detect ground. Next, an optimal walkable direction search algorithm is employed to find the direction that visually impaired people can follow. This direction will then be converted to beep sound to give the users instructions. When the user wants to know the surroundings, he can double tap the smartphone screen to trigger the 2.5-D object detection function. The surrounding information including category, distance and orientation of all obstacles will be provided to the user through speech. The algorithms will be described in the following sections in detail.

\subsection{Time-Dependent Adaptive Ground Detection}\label{sec:Ground_detect}
\begin{figure}[htbp]
	\begin{center}
		\includegraphics[scale=0.35]{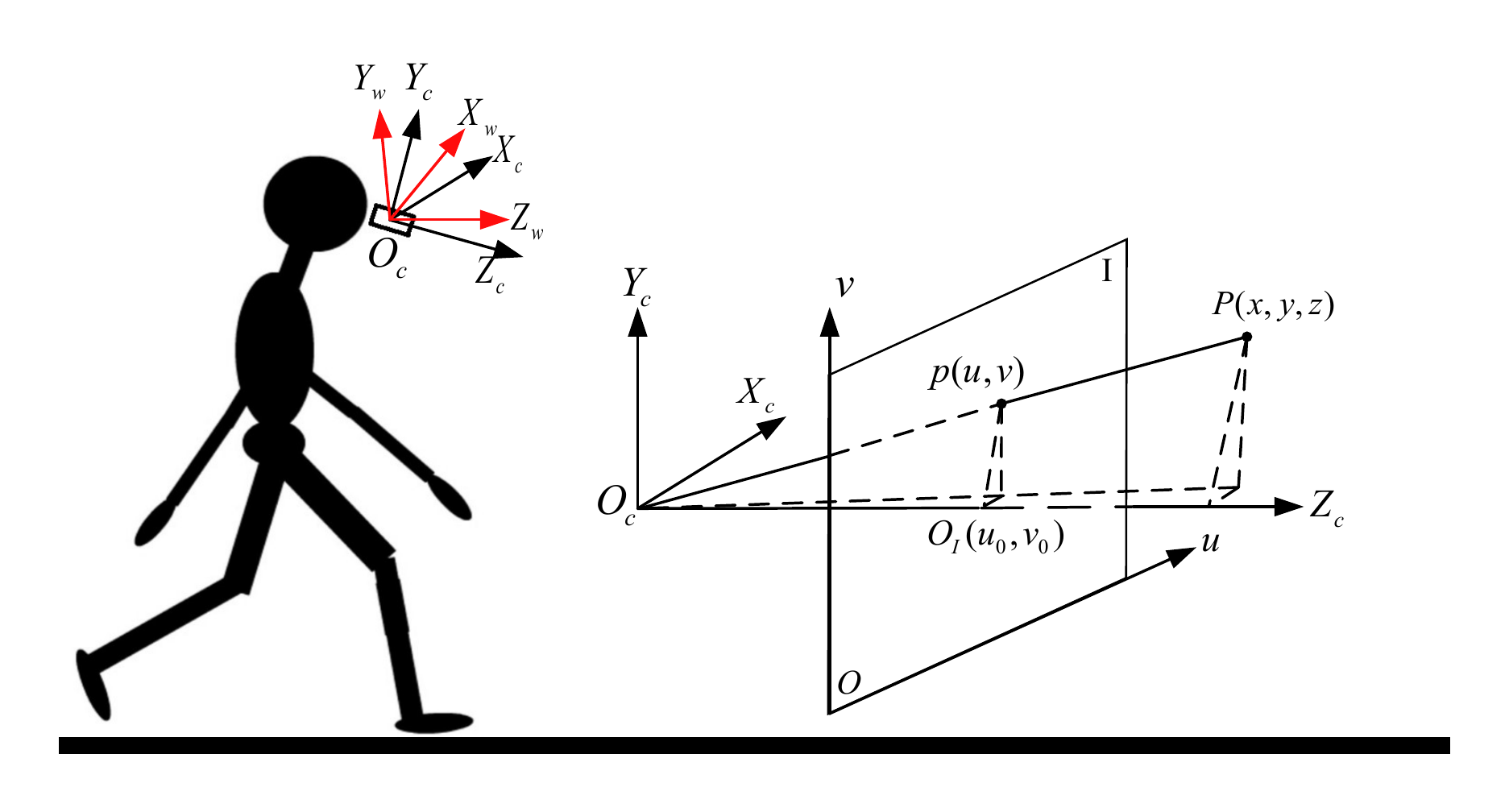}
	\end{center}
	\caption{Coordinate system transformation.}
	\label{fig:3}
\end{figure}

\textbf{Pointcloud reconstruction:}
As shown in Fig. \ref{fig:3}, the camera coordinate system $ \boldsymbol{X}_c\boldsymbol{Y}_c\boldsymbol{Z}_c $ is centered at the camera, and the positive $\boldsymbol{Z}_c-$axis, $\boldsymbol{Y}_c-$axis, and $\boldsymbol{X}_c-$axis are defined as the camera's facing direction, up direction and left direction respectively. The world coordinate system $ \boldsymbol{X}_w\boldsymbol{Y}_w\boldsymbol{Z}_w $ is centered at the camera coordinate system center, the positive $\boldsymbol{Z}_w-$axis, $\boldsymbol{Y}_w-$axis, and $\boldsymbol{X}_w-$axis are the user's facing direction, vertically upward direction, and left direction respectively. Both coordinate systems are the left-handed Cartesian coordinate systems. The pixel value of point $ p(u,v) $ in the depth image represents the distance between point $ P(x,y,z) $ and the camera, which is equal to  $ z $. With the camera attitude angle measured by the IMU, the corresponding 3-D pointcloud in the world coordinate system can be calculated through:
\begin{equation}\label{eqn:1}
\begin{aligned}
\left\{ \begin{array}{l}
\left[\begin{array}{l}
{x_w}\\
{y_w}\\
{z_w}
\end{array} \right] = z{\bf{E}}{{\bf{K}}^{ - 1}}\left[{\begin{array}{*{20}{c}}
	u\\
	v\\
	1
	\end{array}} \right]\\[5mm]
{\bf{E}} = \left[ {\begin{array}{*{20}{c}}
	{\cos \gamma }&{ - \sin \gamma }&0\\
	{\sin \gamma }&{\cos \gamma }&0\\
	0&0&1
	\end{array}} \right]\left[ {\begin{array}{*{20}{c}}
	1&0&0\\
	0&{\cos \alpha }&{ - \sin \alpha }\\
	0&{\sin \alpha }&{\cos \alpha }
	\end{array}} \right]\\[5mm]
{\bf{K}} = \left[ {\begin{array}{*{20}{c}}
	{{f_x}}&0&{{u_0}}\\
	0&{{f_y}}&{{v_0}}\\
	0&0&1
	\end{array}} \right]
\end{array} \right.,
\end{aligned}
\end{equation}
where $K$ is the camera intrinsic parameter matrix, point $ (x_w,y_w,z_w) $ is the reconstructed point in the world coordinate system, $\alpha$ is the camera pitch angle, $\gamma$ is the camera roll angle.

\begin{figure}[htbp]
	\begin{center}
		\includegraphics[width=0.95\linewidth]{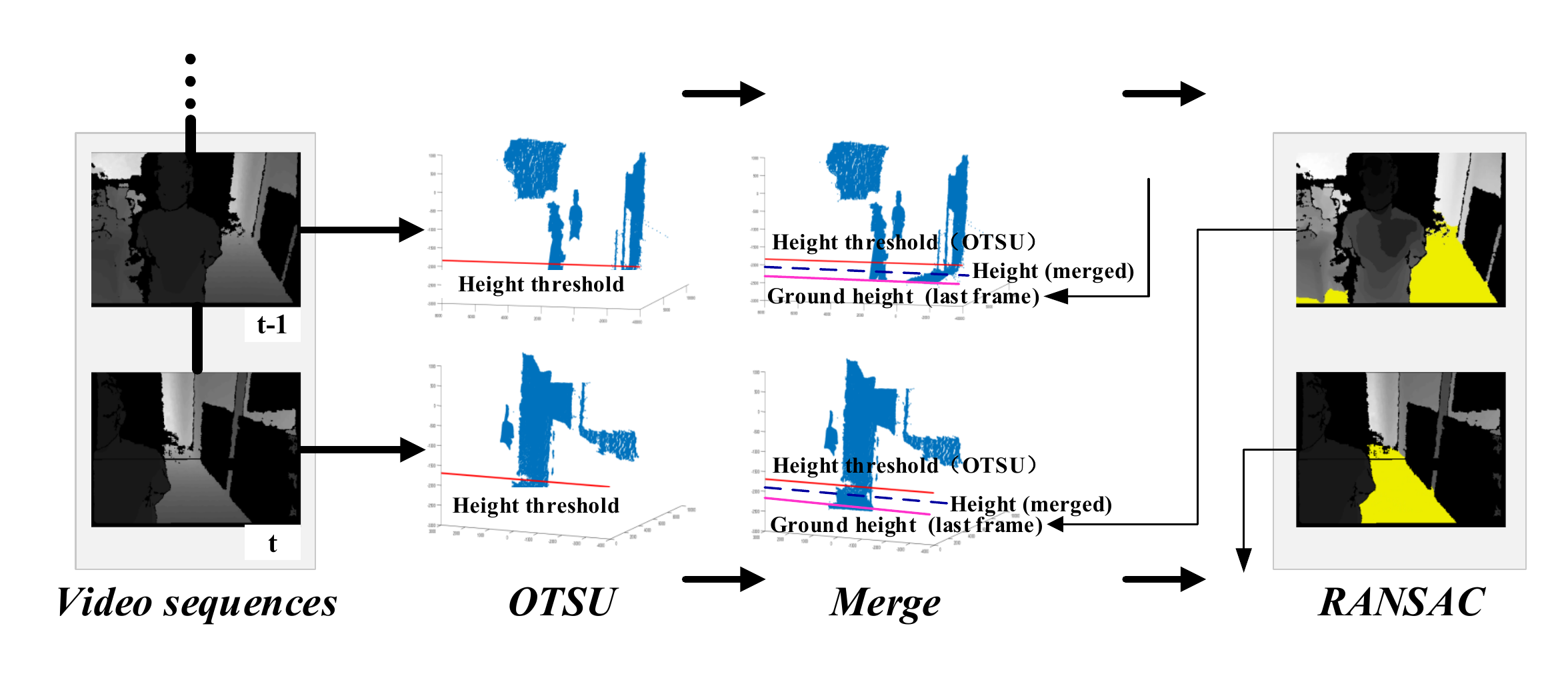}
	\end{center}
	\caption{Ground fitting.}
	\label{fig:4}
\end{figure}

\textbf{Coarse ground fitting:}
As shown in Fig. \ref{fig:4}, the initial ground height threshold $T{Y_{roi}}$ is calculated adaptively using the OTSU algorithm \cite{31} in the current frame. Since the change of ground height in two adjacent frames is usually limited, the ground height  $T{Y_{pre}}$ of the previous frame is used to reduce the perturbation of other planes (e.g. desk, sofa) (see Fig. \ref{fig:9}). The final ground height threshold is computed as:
\begin{equation}\label{eqn:2}
\begin{aligned}
TY = \lambda T{Y_{roi}} + \mu T{Y_{pre}},
\end{aligned}
\end{equation}
where $\lambda,\mu$ are the weights.

Due to the inherent limitation of the depth camera, the depth accuracy always drops down with the increase of distance. Besides, the obstacles that are too far away from the person do not need to be considered. Therefore, only the points within a threshold $TZ$ are used in order to reduce computation cost. By making use of the ground height $TY$ and the distance threshold $TZ$, the 3-D points for fitting the coarse ground can be computed as:
\begin{equation}\label{eqn:3}
\begin{aligned}
{F_{init}} = \{ p(x,y,z)|y < TY,0 < z < TZ\} .
\end{aligned}
\end{equation}
Then the coarse ground is fitted with RANSAC algorithm \cite{15}, and represented as:
\begin{equation}\label{eqn:4}
\begin{aligned}
Ax + By + Cz + D = 0.
\end{aligned}
\end{equation}

\textbf{Ground refinement:}
The normal vector ${\vec n_{ground}}$ of coarse ground plane can be obtained directly by Eq. \ref{eqn:4}, and the ground pitch angle $\phi$ will then be computed through:
\begin{equation}\label{eqn:5}
\begin{aligned}
\phi  = \arccos (\frac{{{{\vec n}_{ground}} \cdot {{\vec n}_{xoz}}}}{{\left| {{{\vec n}_{ground}}} \right|\left| {{{\vec n}_{xoz}}} \right|}}),
\end{aligned}
\end{equation}
where ${\vec n_{xoz}}$ is the normal vector of plane $X_wOZ_w$.

According to the ground pitch angle and the empirical slope angle, the coarse ground will be classified as one of the four types: horizontal, upslope, downslope and non-ground. If it is non-ground, then the visually impaired people will be directly informed that they cannot move on; otherwise, the coarse ground will be refined with the unevenness tolerance $\sigma$ through:
\begin{equation}\label{eqn:6}
\begin{aligned}
\left\{ \begin{array}{l}
F = \{ p(x,y,z)|p \in {F_{init}},dist(p,{F_{init}}) \le \sigma \} \\
dist(p,{F_{init}}) = \frac{{\left| {Ax + By + Cz + D} \right|}}{{\sqrt {{A^2} + {B^2} + {C^2}} }}
\end{array} \right.,
\end{aligned}
\end{equation}
where $dist(p,{F_{init}})$ is the distance from point $p$ to the coarse ground, $F$ is the final 3-D point cloud of refined ground.

Finally, the refined ground height $H$ is obtained through:
\begin{equation}\label{eqn:7}
\begin{aligned}
H = \frac{1}{k}\sum\limits_{i = 1}^k {{y_i}} ,{p_i}({x_i},{y_i},{z_i}) \in F,
\end{aligned}
\end{equation}
and it will be used in next frame.

\subsection{Optimal Walkable Direction Search}
If no ground is detected, the system will directly inform the visually impaired people and stop proceeding the optimal walkable direction search algorithm. Otherwise, the optimal walkable direction is calculated as follows.

Since the walkable direction relies on the detected ground, only the 3-D points within the ground plane need to be considered for walkable direction search. These points are selected by:
\begin{equation}\label{eqn:8}
\begin{aligned}
\begin{array}{l}
P = \{ ({x_i},{y_i},{z_i})|\mathop {\min (x)}\limits_{P(x,y,z) \in F}  \le {x_i} \le \mathop {\max (x)}\limits_{P(x,y,z) \in F} ,\\[3mm]
~~~~~~~~H \le {y_i} \le H + \frac{{\left| D \right|}}{{\sqrt {{A^2} + {B^2} + {C^2}} }} + \varepsilon,\\[3mm]
~~~~~~~~0 \le {z_i} \le \mathop {\max (z)}\limits_{P(x,y,z) \in F} \} .
\end{array}
\end{aligned}
\end{equation}
where $F$ is the 3-D points on ground plane, $H$ is the ground height, $A,B,C,D$ are the plane parameters defined in Eq. \ref{eqn:4}, and $\varepsilon $ is a constant for preventing the person from being collided with overhanging obstacles.

\begin{figure}[b]
\begin{center}
	\includegraphics[width=0.9\linewidth]{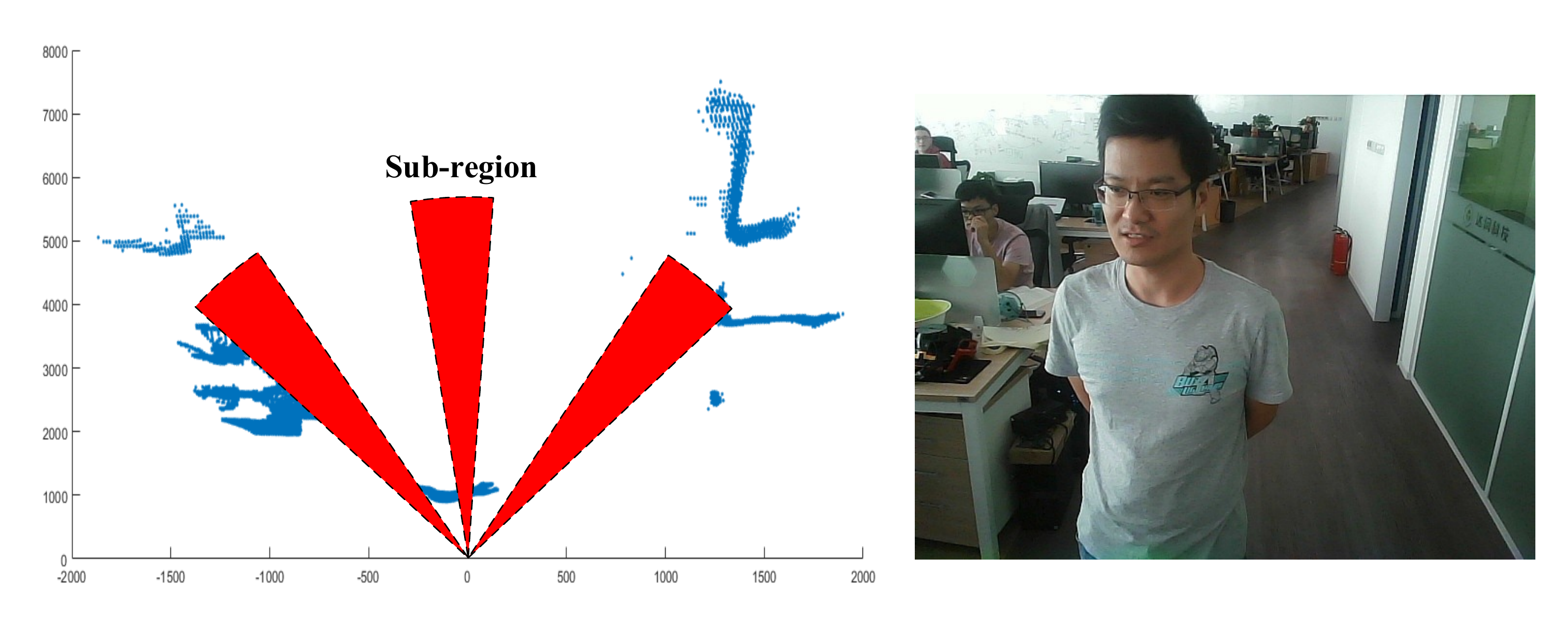}
\end{center}
\caption{Pointcloud that project on the plane $X_wOZ_w$.}
\label{fig:5}
\end{figure}

Then the 3-D points $P$ are projected onto the plane $X_wOZ_w$ (see Fig. \ref{fig:5}). The nearest points in all sectors (each sector is represented as the sub-region in Fig. \ref{fig:5} and the angle of each sub-region is 0.5$^\circ$) can be easily obtained. The award of each sector is computed as:
\begin{equation}\label{eqn:9}
\begin{aligned}
\left\{ \begin{array}{l}
n = \frac{{180}}{\pi }*\frac{{\arcsin {{{w_{sw}}} \mathord{\left/
				{\vphantom {{{w_{sw}}} {{z_i}}}} \right.
				\kern-\nulldelimiterspace} {{z_i}}}}}{\theta }\\
award[i] = \alpha \underbrace {(\frac{\pi }{2} - \theta \left| {(i - \frac{N}{2})} \right|)}_{angle~award} + \beta \underbrace {\mathop {\min }\limits_{j = i:i + n} {z_j}}_{distance~award}
\end{array} \right.,
\end{aligned}
\end{equation}
where $\theta$ is the angle of a sector, $w_{sw}$ is the passable width (greater than the person's body width), $z_i$ is the $z-$axis coordinate values of the nearest point in sector $i$, $\alpha$ and $\beta$ are the weights, and $N$ is the total number of sectors. This award function ensures that the person moves toward the direction with smaller turn angle and longer traversable distance.

Next, the optimal walkable direction is obtained by:
\begin{equation}\label{eqn:10}
\begin{aligned}
\gamma=\left\{ {\begin{array}{*{20}{l}}
{None,~if~{z_{\mathop {\arg \max }\limits_i (award[i])}} < \tau}\\
{\theta ({i_{\max }} - \frac{N}{2}),~else}
\end{array}} \right.,
\end{aligned}
\end{equation}
where $\tau $ is a distance threshold, $z_{\mathop {\arg \max }\limits_i (award[i])}$ is the nearest distance of the sector with the maximum award, $i_{\max }$ is the index of the sector with the maximum award, and $N$ is the total number of sectors. If $z_{\mathop {\arg \max }\limits_i (award[i])}$ is less than a small value $\tau $, it is considered to have a very large risk of collision with obstacles. In that case, the optimal walkable direction does not exist, and the system will inform the users to turn left or right with a large angle to search a walkable direction. If $z_{\mathop {\arg \max }\limits_i (award[i])}$ is larger than $\tau $, the optimal walkable direction is that corresponding to the sector with the maximum award. If the turning angle is very small (e.g. $\left| \gamma  \right| \le 5^\circ$), the system will directly inform the users to go straight to prevent them from vacillating to the left and right.

\subsection{2.5-D Object Detection}
\textbf{CNN based object detection:} As the MobileNet V2 \cite{32} achieves relatively better results than other networks \cite{26}\cite{28}\cite{29}\cite{30} by using depth-wise separable convolutions, and has lower computational complexity and smaller model size, it is utilized here for object detection. The training is implemented on the COCO dataset, which includes 91 classes, such as person, car, bus, chair, etc. These classes are enough for helping visually impaired people have a general perception of surroundings. However, 2-D object detection cannot be directly used for blind navigation due to the lack of object distance information. For example, if an object painted on ground is recognized as an obstacle, it will lead visually impaired people to make an incorrect decision. Therefore, we also use the depth image based object detection to solve the above problem.

\textbf{Depth image based object detection:} The detected ground (see Section \ref{sec:Ground_detect}) is firstly removed from the depth image. Then the close morphological processing$\footnote{https://opencv.org/}$ is performed to merge the small objects. Next, the external contours of the obstacles are extracted and their areas are computed. If the area is less than the threshold $S$, the corresponding obstacle will be merged into its nearest obstacle or taken as noise; otherwise, the obstacle location can be obtained as:

\begin{enumerate}
\item Compute the moment of each contour, and the centroid of the contour can be obtained according to the zero-order and the first-order moment;
\item According to the contour centroid and the camera intrinsic parameter matrix $K$ in Eq. \ref{eqn:1}, the obstacle location is represented as $({\theta _{pitch}},{\theta _{yaw}},z)$:
\begin{equation}\label{eqn:11}
\begin{aligned}
\left\{ \begin{array}{l}
{\theta _{pitch}} = \frac{{180}}{\pi }\arctan (\frac{{center.x - {u_0}}}{{{f_x}}})\\
{\theta _{yaw}} = \frac{{180}}{\pi }\arctan (\frac{{center.y - {v_0}}}{{{f_y}}})\\
z = {z_{center}}
\end{array} \right.,
\end{aligned}
\end{equation}
where $center(x,y)$ is the contour centroid, and $z_{center}$ is the depth value of contour centroid.
\end{enumerate}

\begin{figure}[htbp]
	\begin{center}
		\includegraphics[width=0.95\linewidth]{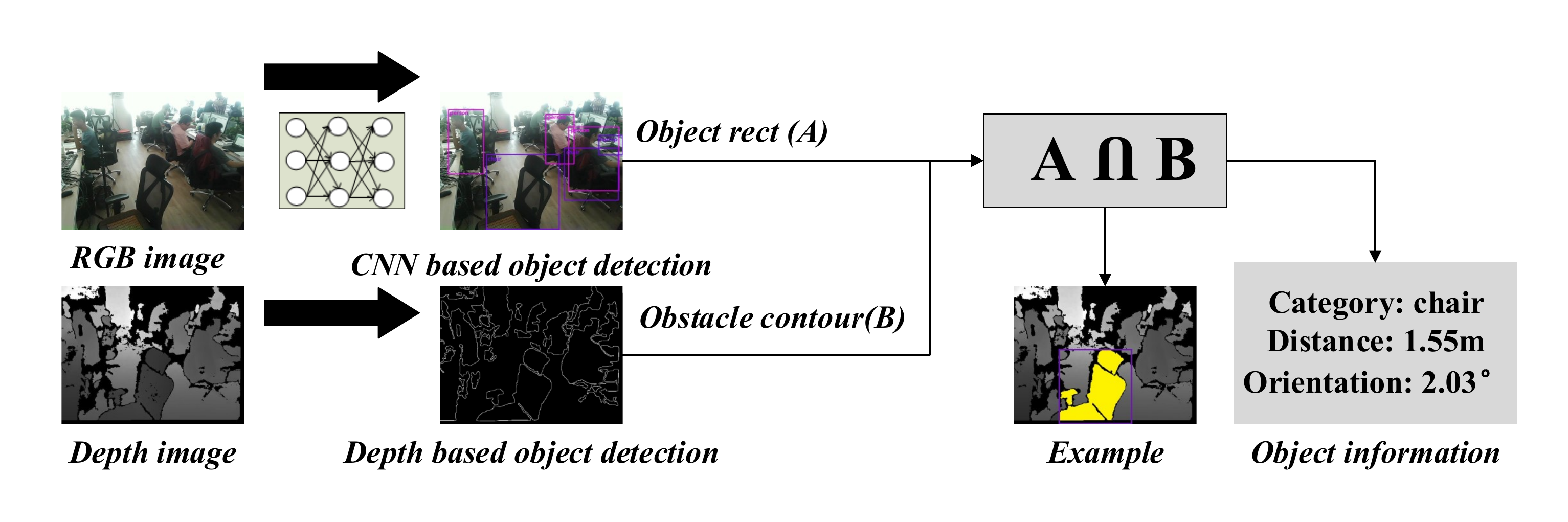}
	\end{center}
	\caption{2.5-D object detection (the yellow region (i.e. $C$ in Eq. \ref{eqn:12})).}
	\label{fig:6}
\end{figure}

\textbf{Combination:} The objects obtained by MobileNet V2 in the RGB image can be easily mapped to the depth image according to camera calibration matrix $[R,t]$. Then the intersection area $C$ between the mapped area $A$ and the detected contour $B$ can be calculated (see Fig. \ref{fig:6}):
\begin{equation}\label{eqn:12}
\begin{aligned}
C = A \cap B.
\end{aligned}
\end{equation}
If $\frac{S_C}{{\max (S_A,S_B)}}$ ($S_i$ represents the area of region $i$) is greater than a threshold $\zeta$ (e.g. 0.7), it means the mapped area and the detected contour is the same object. Then the object distance is the minimum non-zero depth value within the intersection area $C$, and the object orientation relative to the user can be obtained with Eq. \ref{eqn:11}. This way, the key surrounding information including obstacle category, distance and orientation can be provided.

\subsection{Audio Feedback}
\textbf{Navigation feedback:} The beep sound is used to provide the navigation information to the user. When the visually impaired person encounters an obstacle that blocks his way, the system will keep beeping to alarm him that he cannot go straight. In that case, he should turn left or right to search the optimal walkable direction. When the beep sound stops, the user can continue moving.

\begin{figure}[htbp]
	\begin{center}
		\includegraphics[scale=0.6]{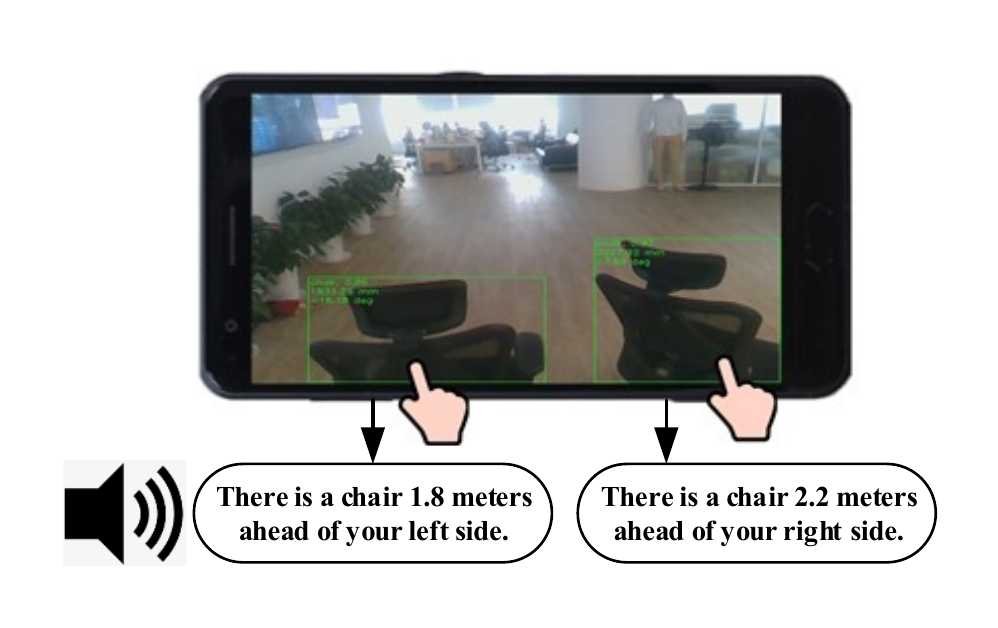}
	\end{center}
	\caption{2.5-D object detection feedback.}
	\label{fig:7}
\end{figure}

\textbf{2.5-D object detection feedback:} When the visually impaired person walks in a relatively complicated environment, the 2.5-D object detection function can be activated by double tapping the smartphone screen, and the results will be converted to speech via a text-to-speech module and broadcast to the user. An example of 2.5-D object detection feedback is shown in Fig. \ref{fig:7}, the object category (e.g. chair), distance (e.g. 1.8 m) and orientation (e.g. ahead of left side) are displayed. Because the visually impaired people are usually insensitive to the exact azimuth angle of the object, only 3 direction types: left-front ($<-5^\circ$), front ($\ge-5^\circ\&\le5^\circ$), and right-front ($>5^\circ$) are provided.

\section{Experimental Results and Discussions}
The performance of the proposed system is evaluated in real scenarios. Firstly, the ground detection is tested since it plays an important role in the whole system. Then, we recruit 20 visually impaired persons to test the navigation and object detection function. We follow the protocol approved by the Bejing Fangshan District Disabled Persons' Federation for recruitment and experiments. All the people who participated in this experiment approved that the results (including data, images and videos) can be published with anonymity. Finally, the computational cost is evaluated to test the real-time performance.

\subsection{Experiments on Ground Detection}
\begin{figure}[htbp]
	\begin{center}
		\includegraphics[scale=0.6]{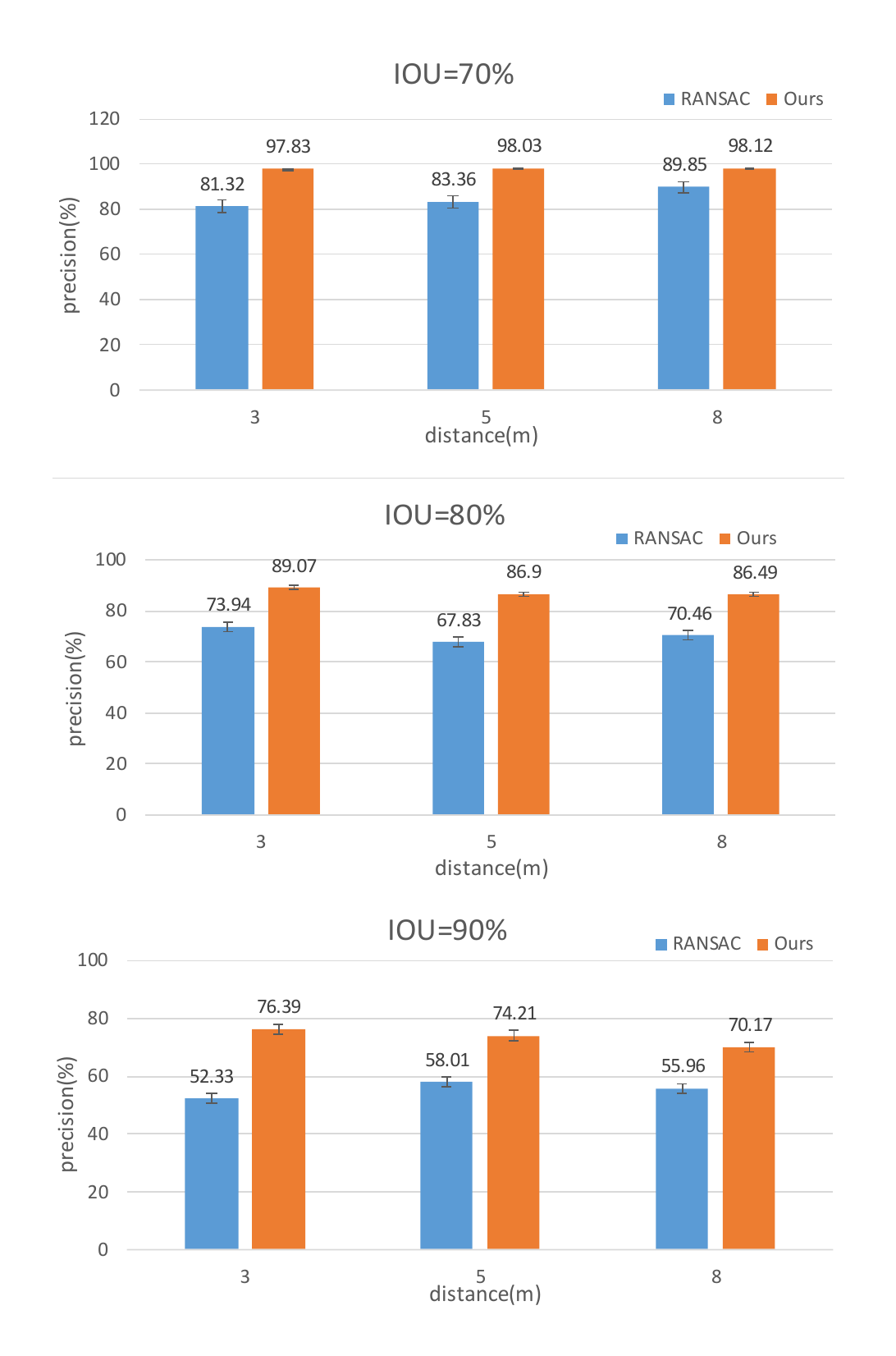}
	\end{center}
	\caption{Precision of ground in different distance and different IOU.}
	\label{fig:8}
\end{figure}

To evaluate the proposed ground detection method quantitatively, we manually labelled a random sample of 1000 images captured in indoor scenario to get the ground truth. Fig. \ref{fig:8} shows the precision of the detected ground with different distance and different Intersection over Union (IOU) percentages. The IOU percentage is computed as:

\begin{equation}\label{eqn:13}
\begin{aligned}
IOU = \frac{{{N_\cap }}}{{{N_{{detected}}} + {N_{{ground truth}}}}},
\end{aligned}
\end{equation}
where $N_ \cap$ is the number of overlapping pixels between the detected ground and the ground truth, $N_{detected}$ is the number of detected ground pixels, and $N_{ground truth}$ is the number of ground truth pixels.

\begin{figure}[htbp]
\begin{center}
	\includegraphics[scale=0.2]{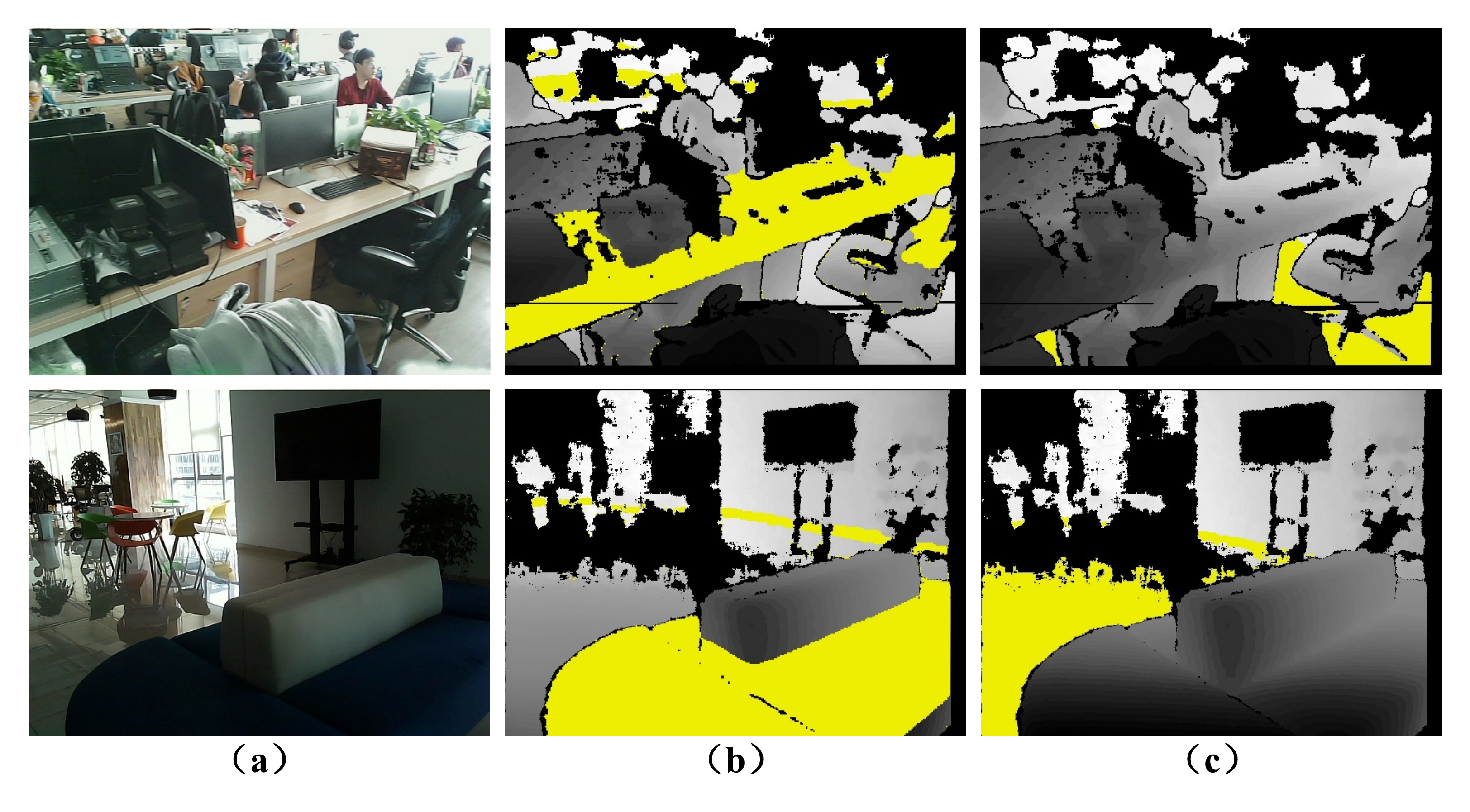}
\end{center}
\caption{Ground detection.(a) the color image (b) the ground (the yellow region) detected by RANSAC algorithm (c) the ground (the yellow region) detected by the proposed algorithm.}
\label{fig:9}
\end{figure}

As shown in Fig. \ref{fig:8}, comparing with the RANSAC algorithm, the proposed method has a higher precision in all scenarios. Some comparative examples are shown in Fig. \ref{fig:9}. By taking the advantage of the ground height continuity among adjacent frames, the proposed method is more robust to the interferences of other planes (such as sofa, table) than the RANSAC algorithm. It provides a solid base for the following object detection.

\subsection{Experiments on Real Scenarios}
\begin{figure}[htbp]
	\begin{center}
		\includegraphics[width=0.9\linewidth]{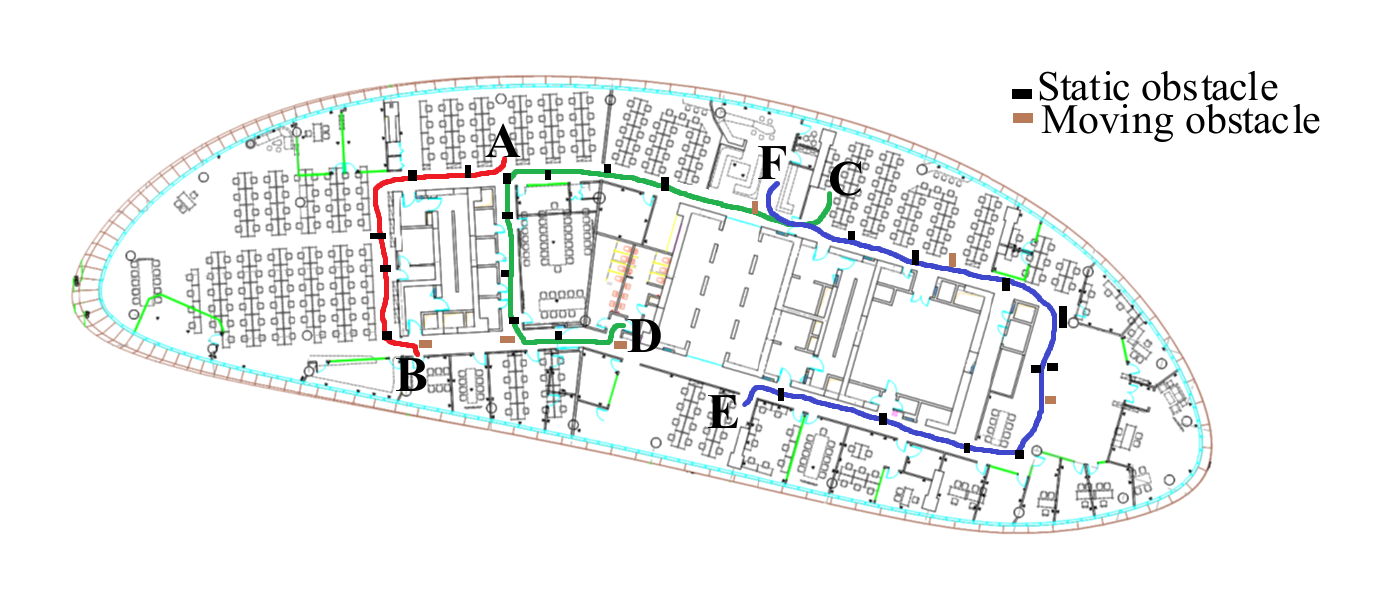}
	\end{center}
	\caption{Test paths.}
	\label{fig:10}
\end{figure}

\textbf{Navigation task:} Three paths in an office (see Fig. \ref{fig:10}) are selected to evaluate the navigation performance. There are many obstacles including static and moving obstacles in these paths. Then 20 visually impaired people who were not familiar with these environments are asked to navigate following these paths with the help of either a white cane or the proposed device. They are trained for about 10 minutes to get familiar with our system. The average walking time and number of collisions with obstacles are recorded (see TABLE \ref{table:1}). The visually impaired people using our proposed system spend less walking time than using the white cane. This proves that the proposed system has a higher navigation efficiency in unfamiliar environments.

\begin{table}[htbp]
	\caption{Results of navigation with Cane and our assistance.}
	\begin{center}
		\begin{tabular}{|c|c|c|c|c|c|c|}
			\hline
			\multicolumn{1}{|c|}{\multirow{2}{*}{Paths}} &
			\multicolumn{1}{c|}{\multirow{2}{*}{Length(m)}} &
			\multicolumn{1}{c|}{\multirow{2}{*}{Obstacles}} &
			\multicolumn{2}{c|}{Average time(s)} & \multicolumn{2}{c|}{Total collisions} \\
			\cline{4-7}  & & &Ours & Cane & Ours & Cane \\
			\hline A$\to$B &~22 & 6 &70.8 &71.5 &0 & 0\\
			\hline C$\to$D &~43 & 11 &232.3 &268.2 &0 & 19\\
			\hline E$\to$F &~80 & 12 &478.1 &512.3 &0 & 23\\
			\hline
		\end{tabular}
		\label{table:1} 
	\end{center}
\end{table}

\begin{figure}[htbp]
	\begin{center}
		\includegraphics[scale=0.25]{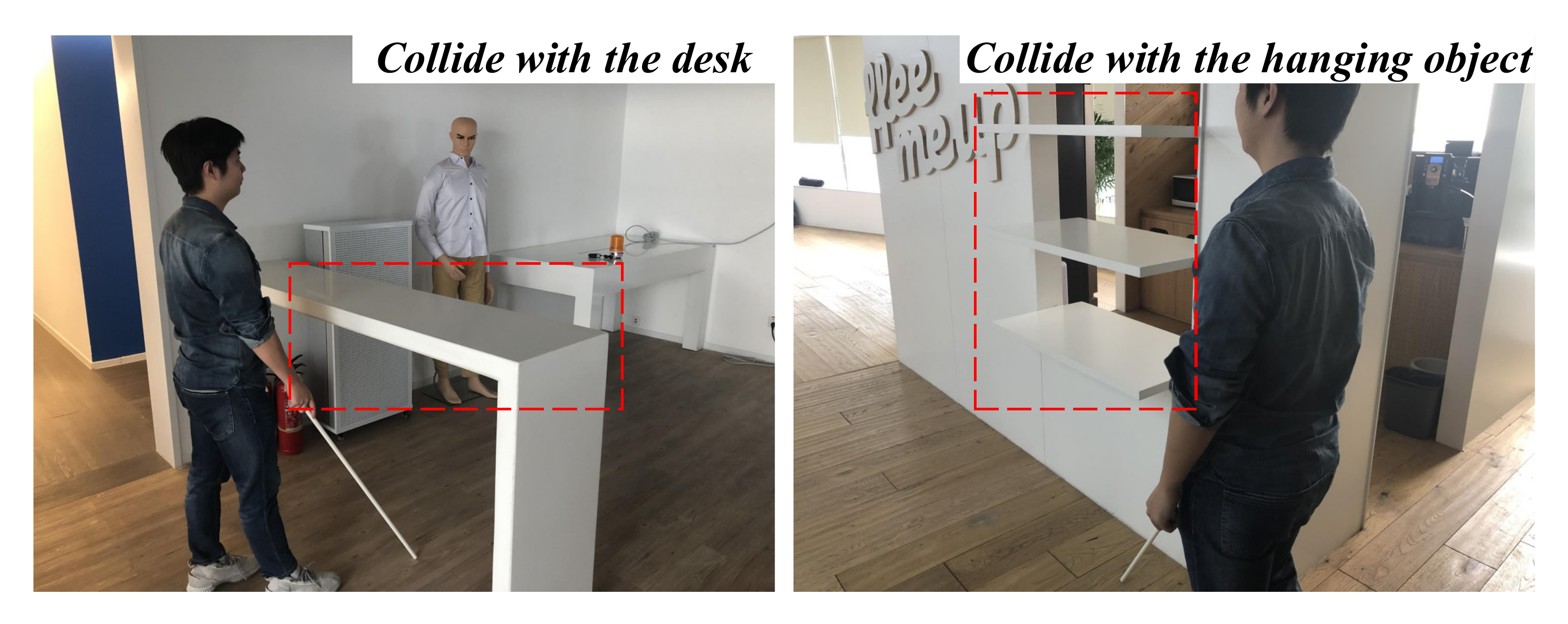}
	\end{center}
	\caption{Collisions when the visually impaired people use a white cane.}
	\label{fig:11}
\end{figure}

As shown in TABLE \ref{table:1}, the users have more collisions with obstacles when using a white cane. This is because they usually use the white cane to detect the obstacles on the ground instead of those hanging in the mid-air, and it is observed that almost all studied users collide with the desk or the hanging objects on the path C$\to$D and E$\to$F (see Fig. \ref{fig:11}). Whereas with the proposed device, they are able to avoid collisions with those obstalces. This proved that the proposed system is more secure for visually impaired people.

\begin{figure}[htbp]
	\begin{center}
		\includegraphics[scale=0.4]{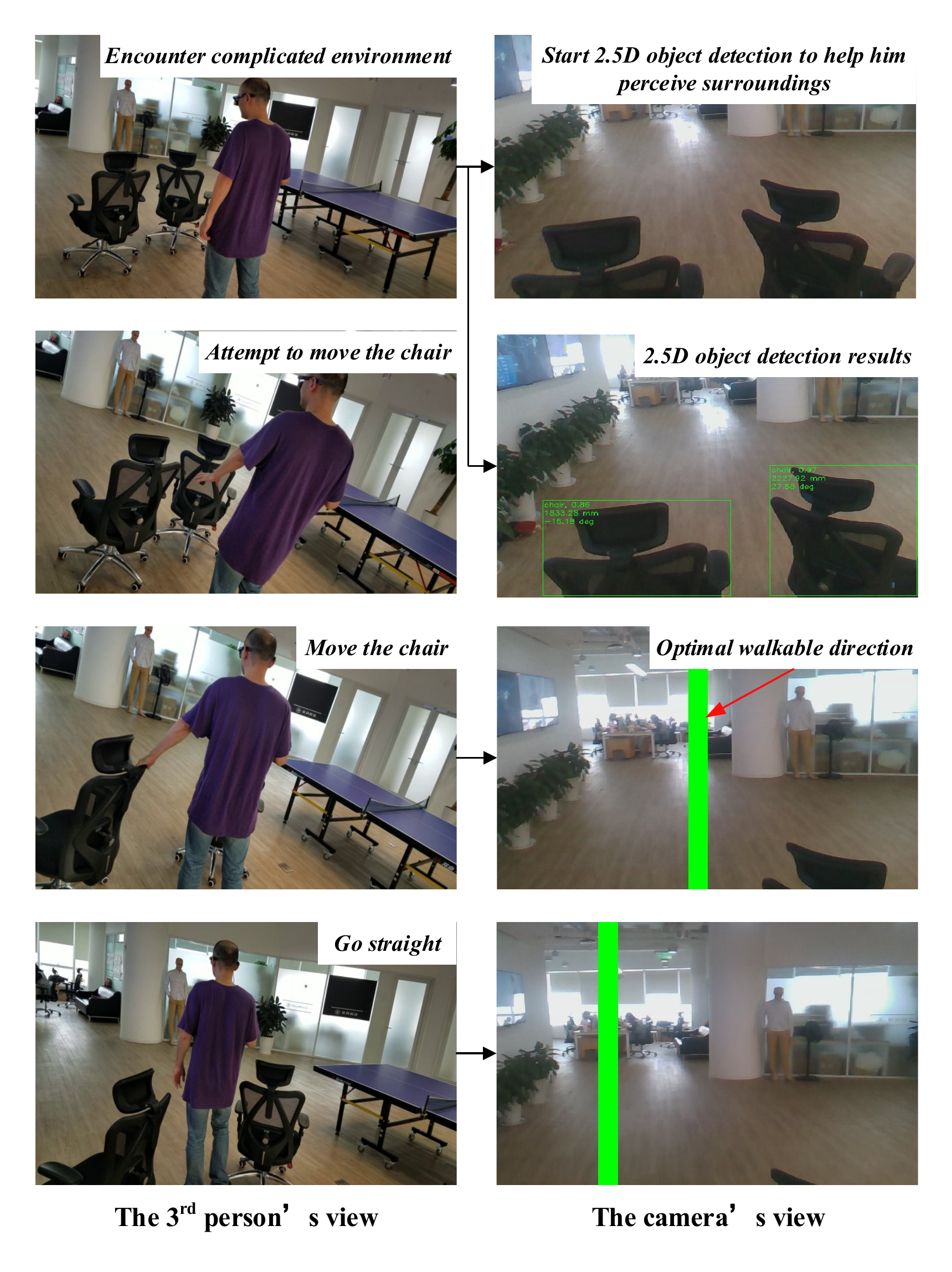}
	\end{center}
	\caption{Example of 2.5-D object detection for helping visually impaired people perceive their surroundings.}
	\label{fig:12}
\end{figure}

\textbf{2.5-D object detection task:} Some complicated scenarios are designed to test if the 2.5-D object detection can help visually impaired people to travel more efficiently. Such examples include the scenario where the way is blocked by two chairs (see Fig. \ref{fig:12}) and crowded with other obstacles. With only the navigation function in our proposed system, the user cannot find the moving direction after multiple searches and will turn back. However, if he activates the 2.5-D object detection and perceives more information about surroundings, he is able to move the chair and search the moving direction. This prevents the visually impaired people from taking more detours and improves their environmental perception abilities. This shows that the 2.5-D object detection indeed helps visually impaired people to travel more efficiently and brings a better travel experience to them.

\subsection{Computational Cost}
All algorithms are implemented on a smartphone with Qualcomm Snapdragon 820 CPU 2.0 GHz and RAM of 4 GB. The average computational time of the proposed system is calculated and shown in TABLE \ref{table:2}. The images acquisition, ground detection, optimal walkable direction search and 2.5-D object detection cost about 0.66 ms, 13.53 ms, 7.19 ms and 114.13 ms respectively. The total time that all algorithms cost excluding the 2.5-D object detection is about 27.17 ms on average. Since the 2.5-D object detection is only activated when the visually impaired people want to know the surrounding information or enter a complicated scenario, the proposed system is able to provide real-time assistance for users' daily traveling.

\begin{table}[htbp]
	\caption{Computational time of the proposed system.}
	\begin{center}
		\begin{tabular}{|c|c|}
			\hline{Processing step} & {Average time (ms)} \\
			\hline RGB and Depth images acquisition & 0.66\\
			\hline Ground detection & 13.53\\
			\hline Optimal walkable direction search & 7.19\\
			\hline 2.5-D Object detection & 114.13\\
			\hline Total (except 2.5-D object detection)	& 22.17\\
			\hline
		\end{tabular}
		\label{table:2} 
	\end{center}
\end{table}

\section{Conclusion}
This paper presents a wearable device which provides navigation and object detection assistance for visually impaired people. To provide reliable navigation, a ground detection algorithm that uses the ground height continuity between two adjacent frames is presented. Then an optimal walkable direction search method is developed to determine the moving direction. To improve the environmental perception ability of visually impaired people, a 2.5-D object detection function is presented. Audio feedback is used to inform the visually impaired people for both the navigation instructions and object detection results. Experimental results show that the proposed system can help visually impaired people travel efficiently and brings a better traveling experience for them. In future, the small-size obstacle detection will be considered.

\end{document}